\documentclass[letterpaper]{article} % DO NOT CHANGE THIS
\usepackage{aaai2026}  % DO NOT CHANGE THIS
\usepackage{times}  % DO NOT CHANGE THIS
\usepackage{helvet}  % DO NOT CHANGE THIS
\usepackage{courier}  % DO NOT CHANGE THIS
\usepackage[hyphens]{url}  % DO NOT CHANGE THIS
\usepackage{graphicx} % DO NOT CHANGE THIS
\usepackage{natbib}  % DO NOT CHANGE THIS AND DO NOT ADD ANY OPTIONS TO IT
\usepackage{caption} % DO NOT CHANGE THIS AND DO NOT ADD ANY OPTIONS TO IT
\usepackage{algorithm}
\usepackage{algorithmic}

\usepackage{newfloat}
\usepackage{listings}
\usepackage[misc]{ifsym} 
\def\ie{\emph{i.e.}}
\def\eg{\emph{e.g.}}

\def\sota{{\em state-of-the-art~}} 
\def\HCE{{\rm HCE}}
\usepackage{graphicx}
\usepackage{amsmath}
\usepackage{amssymb}
\usepackage{multirow}
\usepackage{makecell}
\usepackage{bbding}
\usepackage{makecell}
\usepackage{makecell}
\usepackage{xspace}
\usepackage{booktabs}
\usepackage{nicematrix} 
\DeclareCaptionStyle{ruled}{labelfont=normalfont,labelsep=colon,strut=off} % DO NOT CHANGE THIS
\floatstyle{ruled}
\newfloat{listing}{tb}{lst}{}
\floatname{listing}{Listing}
\title{Toward Real-World High-Precision Image Matting
and Segmentation}
\author {
    Haipeng (Rydeen) Zhou\textsuperscript{\rm 1},
    Zhaohu Xing\textsuperscript{\rm 1},
    Hongqiu Wang\textsuperscript{\rm 1},
    Jun Ma\textsuperscript{\rm 1,2},
    Ping Li\textsuperscript{\rm 3},
    Lei Zhu\textsuperscript{\rm 1,2}\thanks{Corresponding author: Lei Zhu (leizhu@hkust-gz.edu.cn) .}
}
\affiliations {
    \textsuperscript{\rm 1}The Hong Kong University of Science and Technology (Guangzhou)\\
    \textsuperscript{\rm 2}The Hong Kong University of Science and Technology\\
    \textsuperscript{\rm 3}The Hong Kong Polytechnic University\\
}

\usepackage{bibentry}
\begin{document}

\maketitle

\begin{abstract}
		High-precision scene parsing tasks, including image matting and dichotomous segmentation, aim to accurately predict masks with extremely fine details (such as hair). Most existing methods focus on salient, single foreground objects. While interactive methods allow for target adjustment, their class-agnostic design restricts generalization across different categories. Furthermore, the scarcity of high-quality annotation has led to a reliance on inharmonious synthetic data, resulting in poor generalization to real-world scenarios.
		To this end, we propose a Foreground Consistent Learning model, dubbed as FCLM, to address the aforementioned issues. Specifically, we first introduce a Depth-Aware Distillation strategy where we transfer the depth-related knowledge for better foreground representation. Considering the data dilemma, we term the processing of synthetic data as domain adaptation problem where we propose a domain-invariant learning strategy to focus on foreground learning. 
		To support interactive prediction, we contribute an Object-Oriented Decoder that can receive both visual and language prompts to predict the referring target.
		Experimental results show that our method quantitatively and qualitatively outperforms \sota methods.
	\end{abstract}

\begin{links}
    \link{Code}{https://github.com/haipengzhou856/FCLM}
\end{links}

\section{Introduction}    
	High-accuracy image matting and segmentation aim to predict fine-grained masks, which are essential for vision tasks such as AR/VR, image editing, and so on~\cite{goferman2011context,liu2021fully}. Compared to conventional dense prediction, this task requires significantly finer details, making accurate boundary localization and semantic consistency  challenging.

	Data matters, yet real-world fine-grained annotations are extremely costly. Due to the difficulty of labeling, a large portion of existing matting datasets~\cite{li2023referring,xu2017deep,hou2019context} are synthetic and inharmonious.
	Although synthetic data can enhance the training effect~\cite{qian2024maskfactory}, this is based on the premise that the distribution of the images is normal, rather than simply stitching the foreground with different backgrounds to create jarring images. 
	The reliance on inharmonious synthetic data in image matting introduces visual inconsistencies and domain gaps, undermining model generalization. To demonstrate this problem, we reproduce cutting-edge methods on the P3M dataset~\cite{li2021privacy}, replacing its original backgrounds with BG20K images~\cite{li2021privacy} to create inharmonious synthetic training data. 
    As shown in Tab.~\ref{tab:1}, models trained on synthetic data exhibit a significant performance drop when evaluated on real-world images. This discrepancy highlights the limitations of inharmonious data for training.
    However, considering the difficulty in annotation, using such inharmonious data in matting scenarios is a data augmentation measure adopted out of necessity. Our findings underscore the critical need for techniques to properly utilize synthetic data and unleash their power.
    \begin{table}[t]
		\centering
		\setlength{\tabcolsep}{1.5mm}
		%\resizebox{\textwidth}{!}{
			\scriptsize
			\begin{tabular}{c|cccc}
				\Xhline{1.02pt}
				\multicolumn{1}{c|} {Methods} &\multicolumn{1}{c}{SAD$\downarrow$} &\multicolumn{1}{c}{MSE$\downarrow$}&\multicolumn{1}{c}{Grad$\downarrow$}&\multicolumn{1}{c}{Conn$\downarrow$}\\
				\hline
				P3M-Net~\cite{li2021privacy}&8.73&0.0027&13.83&9.14\\
				P3M-Net$\dagger$~\cite{li2021privacy}&12.33&0.0051&16.44&11.67\\
				
				MatteFormer~\cite{park2022matteformer}&7.65&0.0020&12.34&7.89\\
				MatteFormer$\dagger$~\cite{park2022matteformer}&10.84&0.0064&15.67&10.44\\
				
				MODNet~\cite{ke2022modnet}&10.33&0.0057&15.35&14.35\\
				MODNet$\dagger$~\cite{ke2022modnet}&23.41&0.0154&26.38&18.95\\
				\Xhline{1.02pt}
			\end{tabular}
			%}
            		\caption{Comparison of results obtained by training on synthetic ($\dagger$) and real data on P3M-500-P testing set.}
            \label{tab:1}
	\end{table}

	Furthermore, most existing methods are designed to be object-specific (\eg, portrait-focused) or salient-object oriented. The former~\cite{li2021privacy,ke2022modnet,li2022bridging} often suffer from poor generalization, as they are trained and optimized for specific categories and fail on unseen object types. The latter approaches~\cite{yu2024multi,qin2022highly,hu2023high} are category-agnostic, and they struggle to identify which object to segment when multiple salient regions coexist. Neither paradigm supports flexible user interaction or multi-instance segmentation, making them unsuitable for open-set and real-world scenarios. Multi-stage methods~\cite{yao2024matte,li2024matting,ye2024unifying,sun2022human} can alleviate this problem, where an initial coarse mask is first generated and a refine-net is used to polish the prediction. For example, MAM~\cite{li2024matting} integrates Ground-DINO~\cite{liu2024grounding} to generate semantic bounding boxes, and on top of them, SAM~\cite{kirillov2023segment} can produce the initial mask. MAM then proceeds with a refinement stage to improve mask quality. Obviously, such a triple-stage approach is inherently complex. Moreover, the reliance on multiple independent modules could lead to error propagation and reduce overall robustness.
	
	To address the aforementioned challenges, this paper introduces a Foreground Consistent Learning Model (FCLM) for high-accuracy image matting and segmentation. During training, the network takes as input a pair of images that share the same foreground, enabling the model to learn consistent representations across variations in background and context. To enhance spatial reasoning and improve foreground-background separation, we transfer depth-aware priors through a knowledge distillation strategy. Leveraging foreground consistency, we further introduce an adversarial learning loss together with an optimal transport loss to learn domain-invariant features. Additionally, to support flexible user interactions, we propose a prompt encoder that is integrated into the decoding process, enabling object-oriented and semantics-aware predictions.
	
	In sum, our contributions are four-fold:
	\begin{itemize}
		\item We term the inharmonious synthesized images as a domain adaptation problem for better real-world generalization, in which we propose Foreground Consistent Domain Adaptation to encourage the network to focus on the domain invariant features.
		\item We introduce a Depth-Aware Distillation strategy for knowledge transfer, enabling our framework to separate the features between foreground and background.
		
		\item An Object-Oriented Decoder is deployed to support various prompts including visual and text guidance, making our model interactive and semantic-aware.
		
		\item Our overall framework, FCLM, achieves \sota performance on various public datasets, verifying the effectiveness of our approach.
	\end{itemize}
	\section{Related Works}	
	%\textbf{Dense Prediction.}
	\subsection{Dense Prediction}
	Dense prediction requires pixel-level understanding, such as semantic segmentation and depth estimation. Since deep learning era~\cite{he2016deep}, a series of works based on CNN~\cite{shelhamer2016fully,ronneberger2015u,fan2023advances,liu2024primitivenet}, Transformer~\cite{xie2021segformer, wang2024video,ranftl2021vision,jain2023oneformer,zhou2024timeline}, Mamba~\cite{liu2024vmamba,xing2024segmamba,wu2024rainmamba}, or their variants~\cite{chen2021transunet} have emerged.
	 To achieve higher precision in dense prediction, numerous strategies have been developed to enhance the feature representation, like multiscale approaches~\cite{xiao2018unified,zhao2017pyramid}, patch tiling~\cite{li2024patchrefiner}, and cascade structure~\cite{yuan2021hrformer}. 
	 Recently, vision foundation model like SAM~\cite{kirillov2023segment, ravi2024sam} and Depth-Anything~\cite{yang2024depthv1,yang2024depthv2} have shown more powerful performance.
	Despite the significant advancements, the existing general approaches still exhibit a substantial performance disparity when it comes to the editable level, which demands high-precision parsing.
	\subsection{High-Precision Image Parsing}
	%\textbf{High-Precision Image Parsing.}
	High-precision tasks such as image matting and dichotomous segmentation are significantly more challenging which require accurate estimation of fine-grained structures (\eg, hair). 
	Previous image matting approaches~\cite{xu2017deep,hou2019context,li2020natural,liu2021tripartite,park2022matteformer,yao2024vitmatte} typically rely on trimap guidance to predict the alpha matte. 
	However, trimap generation often involves costly manual annotation. Although fully automatic networks~\cite{li2021privacy,ke2022modnet,li2022bridging,qin2022highly,yu2024multi,lin2021real,sun2022human} enable end-to-end inference without user input, their class-agnostic nature tends to produce false positive predictions. 
	Moreover, such architectures could bias the model towards learning salient or isolated single foreground objects, thereby limiting generalization to a broader range of objects. Recently, approaches~\cite{li2024matting,yao2024matte,ke2023segment,liu2024segment} built upon SAM leverage visual prompts to enable more intuitive user interaction, where the coarse masks generated by SAM are further refined. 
	Nevertheless, these methods heavily rely on the quality of the initial mask, and SAM itself lacks sufficient semantic understanding. Although CLIPMat~\cite{li2023referring} introduces referring image matting with textual input, it directly finetunes CLIP for the matting, inevitably suffering from catastrophic forgetting. 
	Moreover, due to CLIP's inherent lack of pixel-level understanding~\cite{zhang2024exploring}, such a straightforward way struggles to achieve satisfactory performance. 
	In contrast, our method can support both visual and textual prompts to predict the specified target accurately, instead of a single object and class-agnostic.

	\begin{figure}[t]
		\centerline{\includegraphics[width=1\columnwidth]{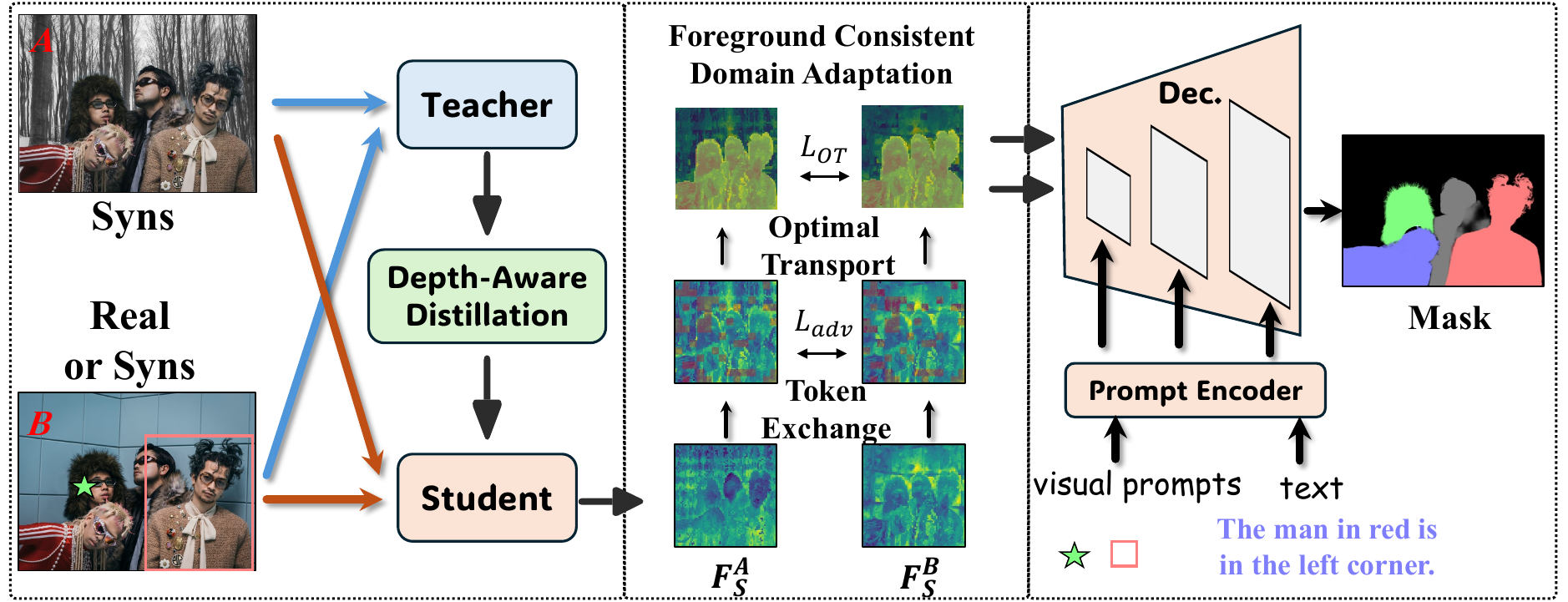}}
		\caption{Overview of the framework of our proposed FCLM. During training, we input two images sharing the same foreground. First, we perform Depth-Aware Distillation to transfer depth-related knowledge from the teacher model to the student model. Then, we apply Foreground Consistent Domain Adaptation to enhance the generalization ability of the model. Our prompt encoder supports various types of prompts, enabling multi-instance and semantic-aware high-precision prediction.}
		\label{fig:method}
	\end{figure}
	%\textbf{Data Reliability.}
	\subsection{Data Reliability}
	Existing real-world datasets focusing on high-precision matting~\cite{li2021privacy,li2022bridging,liu2021tripartite,yu2021mask} or accurate dichotomous segmentation~\cite{qin2022highly} only provide annotations for single objects. 
	Although HIM2K~\cite{sun2022human} contains multiple instances, it is specifically centered on human subjects. 
	A more common design in such datasets is to annotate the foreground object and then composite it with various background images. For example, RefMatte~\cite{li2023referring} introduces a multiple-object matting dataset by attaching diverse foreground instances. 
	However, this simple synthesis-based composition inevitably introduces a domain gap due to the visual inconsistency between foregrounds and backgrounds, leading to suboptimal performance. To address this limitation, we formulate it as a domain adaptation problem, aiming to better generalize from existing datasets.

	\section{Methodology}

		\begin{figure}[t]
	\centering
	%\vspace{-10mm}
	\includegraphics[width=0.9\linewidth]{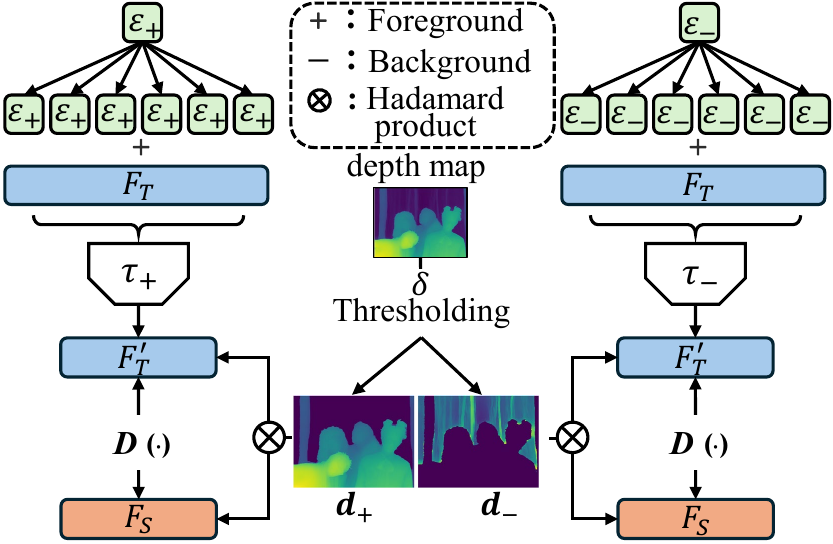}
	\caption{Illustration of the Depth-Aware Distillation. We deploy depth map as guidance to drive foreground and background distillation.}
	\label{fig:DAD}
\end{figure}
	
	\subsection{Overview}
	As depicted in Fig.~\ref{fig:method}, we input the synthesized image $A$ and vanilla (real or synthesized) image $B$ into our pipeline. We first introduce a Depth-Aware Distillation, in which we distill the robustness prior of depth feature to the student model. Next, we apply Foreground Consistent Domain Adaptation to encourage the network to focus on the domain-invariant (\ie, foreground) learning. Finally, our decoder can support various prompts, including visual and text guidance, achieving a target-oriented prediction.
	Note that during the inference stage, only the student model and decoder will be used.

	\subsection{Depth-Aware Distillation}
	In our task, we seek to separate the foreground from the background, which is strongly related to depth estimation. Motivated by this, we propose a Depth-Aware Distillation (DAD) strategy to transfer the depth-related knowledge into a smaller student model. In our practice, we deploy Depth-Anything V2~\cite{yang2024depthv2} as the teacher model while the student model is DINOV2~\cite{oquab2023dinov2}. 
	
	Given the synthesis image $A$ and vanilla image $B$, 
	the student and teacher model will yield encoded features $F_{S}^{A}$, $F_{S}^{B}$,$F_{T}^{A}$, and $F_{T}^{B}$. Instead of aligning the knowledge at the output level which requires auxiliary student heads to predict depth, raising additional computation, we conduct the distillation at the feature level. The standard distillation loss~\cite{hinton2015distilling} can be illustrated as:
	
	\begin{equation}
		\mathcal{L}_{\texttt{kd}} = D(F_{S}^{A}, \tau(F_{T}^{A}))+D(F_{S}^{B},\tau(F_{T}^{B})),
	\end{equation}
	where $D$ is the distance function measuring the discrepancy of the features, and $\tau$ is a project function to align the dimension. However, this simple constraint can introduce feature noises due to the capacity gap~\cite{huang2022knowledge} between teacher and student.

	To better transfer valuable information, we utilize the depth map generated by the teacher model as a guide. As shown in Fig.~\ref{fig:DAD}, inspired by the ViT-Register~\cite{darcet2023vision} we introduce two meta-nets initialized with a context token $\varepsilon$, which are used to project the teacher features into foreground and background representations, respectively. The additional token helps capture global context and reduces artifacts in the feature maps. As a result, the projected feature is produced by $F'_{T} = \tau(F_T+\varepsilon)$, where the dimension of $F'_{T}$ is aligned with the student feature $F_{S}$. To distinguish between foreground and background, we apply an empirical threshold $\delta$ to the depth map, generating corresponding weights  $d_+$ and $d_-$, respectively. Specifically, we set:

\begin{equation}
\begin{split}
    d_+(x, y) &= \begin{cases} \frac{\text{Depth}(x, y)}{\max(\text{Depth})}, & \text{if } \text{Depth}(x, y) > \delta, \\ 0, & \text{otherwise}, \end{cases} \\
    d_-(x, y) &= \begin{cases} \frac{\delta - \text{Depth}(x, y)}{\delta}, & \text{if } \text{Depth}(x, y) \leq \delta, \\ 0, & \text{otherwise}. \end{cases}
\end{split}
\end{equation}
	
	By doing so, we achieve more precise and targeted distillation of relevant features. As a result, the depth-aware distillation for image $A$ can be written as: 
\begin{equation}
\begin{split}
\mathcal{L}_{\texttt{kd}}^{A} &= D(d_{+}^{A}\times F_{S}^{A}, d_{+}^{A}\times \tau_{+}(F_{T}^{A}+\varepsilon_{+}))\\
& \quad + D(d_{-}^{A}\times F_{S}^{A}, d_{-}^{A}\times \tau_{-}(F_{T}^{A}+\varepsilon_{-})).
\end{split}
\end{equation}
	Similarly, we can deploy this formulation for image $B$ to conduct distillation as well.

	%We initialize two meta-net $\tau_{fore}$ and $\tau_{back}$ which used to storing the information of foreground and background. In addition, to ViT-Register

	\subsection{Foreground Consistent Domain Adaptation}
	In this section, we perform consistent learning on features $F^A_S$ and $F^B_S$, which share the same foreground. 
	Inspired by domain adaptation methods that encourage models to focus on domain-invariant components, 
	we refer to our approach as foreground consistent learning, emphasizing its concentration on preserving consistency in the shared foreground across different background images.
	%Inspired by game perspective~\cite{acunadomain}, 
	
		%Note that even when image A and B are both synthesized, their discrepancy between the background can still denote the domains. 
	Following the conventional adversarial learning~\cite{ganin2016domain}, we assign domain labels to input image pairs, \ie, 0 for A and 1 for B to denote their domains. 
	A domain discriminator $h(\cdot)$ is deployed to classify the features  $F^A_S$ and $F^B_S$ into these two domains. Notably, a Gradient Reversal Layer (GRL) \cite{ganin2015unsupervised} is inserted between the student encoder $\mathcal{E}_{S}$ and the discriminator to enforce domain-invariant feature learning. During backpropagation, the GRL will reverse the gradient from $h(\cdot)$ via multiplying the negative factor $-\lambda$. This encourages the student encoder to produce features that confuse the discriminator, thereby reducing the discrepancy between $A$ and $B$. Formally, the adversarial loss for the domain discriminator can be expressed as:
	\begin{equation}
\label{loss:adv}
\begin{split}
    \mathcal{L}_{\texttt{adv}} = \min_{\mathcal{E}_S}\max_{h} \Biggl( &
    \mathbb{E}_{ A} \left[ \log h(\mathcal{E}_S(x^A)) \right] \\
    & + \mathbb{E}_{ B} \left[ \log \left( 1 - h(\mathcal{E}_S(x^B)) \right) \right]
    \Biggr).
\end{split}
\end{equation}

    However, since image $B$ also could be synthesized in a way that fails to confuse the discriminator during adversarial training, it may lead to the collapse of the training process. To mitigate this issue, we introduce a simple yet effective token exchange strategy. Specifically, we propose swapping tokens at identical indices between $F^A_S$ and $F^B_S$. This design offers several intuitive benefits: (1) Exchanged foreground tokens retain contextual coherence due to the presence of the same foreground objects. (2) Exchanging background tokens introduces implicit domain adaptation perturbation signals, helping to suppress domain-specific biases. (3) Shared masks ensure consistent decision boundaries for transitional regions. In practice, we randomly swapping the visual tokens with a ratio of $25\%$.
	
	%In addition, we also devise a Optimal Transport (OT) loss for aligning the domain-invariant distribution for better consistency learning. 
    For better aligning the domain-invariant distribution, we continue to optimize the learning of foreground consistent features.
    Firstly, we deploy the ground truth $\mathcal{M}$ to filter the valid foreground tokens by:
	\begin{equation}
\begin{split}
    F_{fg}&=\{F_{S}[i]~|~\mathcal{M}_{\texttt{patch}}[i]>0\},
\end{split}
\end{equation}
	where the \texttt{patch} denotes we downsample to patch-level resolution for the mask grid. As a result, we have the foreground tokens $F^{A}_{fg}=\{f_{1}^{A},...,f_{K}^{A}\}$ and $F^{B}_{fg}=\{f_{1}^{B},...,f_{K}^{B}\}$, and $K$ is the number of foreground tokens. Thus, we have two empirical domain distributions:
 \begin{equation}
 	u_{A} = \sum_{i=1}^{K} \frac{1}{K}\delta_{f^{A}_{i}}, ~~~~~
 	u_{B} = \sum_{j=1}^{K} \frac{1}{K}\delta_{f^{B}_{j}},
 \end{equation}
 where $\delta$ is a Dirac delta function centered at the domain feature $f^{A}_{i}$ or $f^{B}_{j}$. To align the distribution, we devise an Optimal Transport (OT) loss to optimize it. As a Kantorovich problem~\cite{kantorovich2006translocation}, OT seeks to find a transport plan $\pi$ that minimizes the total cost $\mathcal{C}$ of moving "mass" from different distributions, which can be written as:
\begin{equation}
    \begin{split}
        OT(u_{A}, u_{B}) &\triangleq \min_{\pi \in \Pi(u_{A}, u_{B})} \langle \pi, \mathcal{C} \rangle_{F}, \\
        \textit{s.t.}\quad \Pi(u_A, u_B) &= \left\{ \pi \in \mathbb{R}^{K \times K}_+ \;\middle|\; \pi \mathbf{1}_K = u_A, \right. \\
        &\qquad \left. \pi^\top \mathbf{1}_K = u_B \right\}
    \end{split}
\end{equation}
where $\langle\cdot,\cdot\rangle_{F}$ is the Frobenius inner product, equivalent to $\sum_{i,j} \pi[i,j] \mathcal{C}[i,j]$, and $\Pi(u_A, u_B)$ is the set of all transport plans $\pi$ whose row and column sums match the distributions $u_A$ and $u_B$, respectively. For cost matrix $\mathcal{C}$, to encourage feature alignment in terms of orientation in the latent space~\cite{cheng2021rethinking}, we use cosine dissimilarity to measure the distance. As a result, we can deploy Sinkhorn-nopp scaling algorithm~\cite{chizat2018scaling,frogner2015learning} to minimize the OT loss:
\begin{equation}
\label{loss:ot}
\begin{split}
    \mathcal{L}_{OT} &= \min_{\pi} \sum_{i,j}\pi[i,j] \mathcal{C}[i,j] \\
    &= \frac{1}{K} \times \left(1-\frac{F^{A}_{fg}[i]\cdot F^{B}_{fg}[j]}{||F^{A}_{fg}[i]||_{2}~||F^{B}_{fg}[j]||_{2}}\right).
\end{split}
\end{equation}
	Regarding the background, we do not perform alignment as it has limited contribution to the final prediction.
	
    \begin{figure}[t]
		\centerline{\includegraphics[width=\columnwidth]{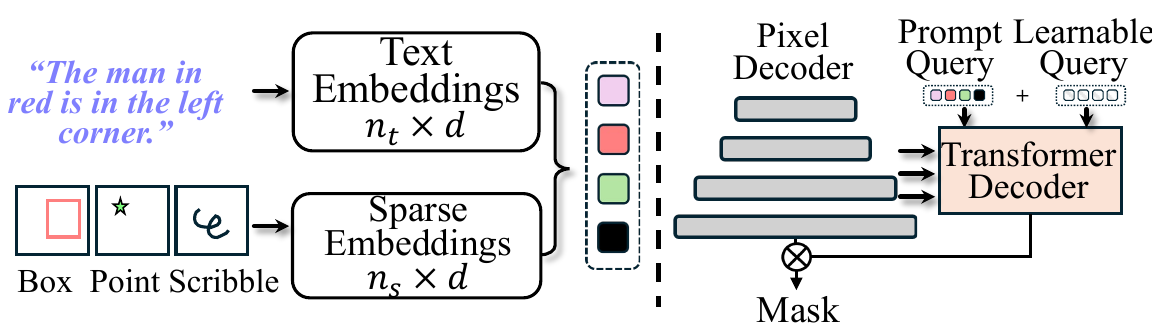}}
		\caption{Illustration of our Object-Oriented Decoder. It supports various prompt inputs to orient the object, achieving interactive high-precision matting or segmentation.}
		\label{fig:decoder}
	\end{figure}

\subsection{Object-Oriented Decoder}
Here, we develop an Object-Oriented Decoder for supporting various prompts. Unlike the complex pipelines~\cite{sun2022human,yao2024matte,li2024matting} which are based on SAM~\cite{kirillov2023segment} or Mask R-CNN~\cite{he2017mask} for coarse mask prediction and subsequent refinement, we adopt the elegant design of SmartMatting~\cite{ye2024unifying} to directly and explicitly prompt the model through visual guidance in an end-to-end manner. Moreover, our framework additionally supports language-based referring settings.
	
	%To start with, our decoder following the design of Mask2Former~\cite{cheng2022masked} which consists of a pixel decoder and transformer decoder.
	To deal with the visual prompt, we embed the coordinates using positional encoding to obtain a sparse embedding. Notably, if no visual prompt is provided, we initialize the corresponding embedding with zeros.
	For the language input, we utilize CLIP's text encoder~\cite{radford2021learning} to generate the text embedding. Previous works~\cite{li2022language,lin2023clip} typically compute similarity between pixel features and the CLIP text embedding for prediction. However, this paradigm is not well-suited to our task, as CLIP lacks fine-grained understanding of image details. Instead, we treat the text embedding as prompt guidance. Furthermore, we introduce an additional learnable query that serves as an implicit context prompt~\cite{zhou2022coop}.
	As illustrated in Fig.~\ref{fig:decoder}, following Mask2Former and SAM~\cite{cheng2022masked,kirillov2023segment}, we employ lightweight pixel decoder and transformer decoder for mask prediction. The concatenated queries enable interaction between the prompts and hierarchical image features. As a result, our pipeline effectively facilitates both prompt integration and accurate prediction.
	
\begin{table*}[t]
    \centering
    \renewcommand\arraystretch{0.65}
    \setlength{\tabcolsep}{1.8mm}
    \resizebox{1\textwidth}{!}{
        \begin{tabular}{r|cccc|cccc}
            \toprule[1.1pt]
            \multicolumn{1}{c|}{\textsc{Conifgs}}& \multicolumn{4}{c|}{\textsc{HIM2K-Natural}} & \multicolumn{4}{c}{\textsc{RefMatte-RW}} \\
            \midrule[1.1pt]
            \multicolumn{1}{c|}{Methods} &\multicolumn{1}{c}{IMQ\textsubscript{MSE}$\uparrow$} &\multicolumn{1}{c}{IMQ\textsubscript{MAD}$\uparrow$}&\multicolumn{1}{c}{IMQ\textsubscript{Grad}$\uparrow$}&\multicolumn{1}{c}{IMQ\textsubscript{Conn}$\uparrow$} &\multicolumn{1}{c}{MSE$\downarrow$} &\multicolumn{1}{c}{SAD$\downarrow$}&\multicolumn{1}{c}{Grad$\downarrow$}&\multicolumn{1}{c}{Conn$\downarrow$} \\
            \midrule
            InstMatt~\cite{sun2022human}&81.34&70.26&\underline{74.90}&72.60&$/$&$/$&$/$&$/$\\
            CLIPMat~\cite{li2023referring}&$/$&$/$&$/$&$/$&0.0474&85.83&$/$&$/$\\
            MatAny~\cite{yao2024matte}&75.68&62.67&46.78&68.55&0.0270&52.91&25.17&5.13\\
            MAM~\cite{li2024matting}&81.67&68.78&51.79&72.62&0.0151&29.23&25.85&\textbf{2.83}\\
            SmartMatting~\cite{ye2024unifying} & \underline{82.73} & \underline{71.44} & 68.58 & \underline{74.33} & \underline{0.0120} & \underline{25.60} & \underline{22.62} & 5.31 \\
            \midrule
            Ours & \textbf{83.48} & \textbf{73.45} & \textbf{75.69} & \textbf{74.83} & \textbf{0.010} & \textbf{21.31} & \textbf{19.38} & \underline{3.33}\\
            \bottomrule[1.1pt]
        \end{tabular}
        
    }
        \caption{Quantitative comparisons on real-world multi-object matting datasets.
        The best ones are highlighted with \textbf{bold} and runner-ups are \underline{underlined}.}
    \label{tab:matte}
\end{table*}

	\begin{figure}[t]\centerline{\includegraphics[width=\columnwidth]{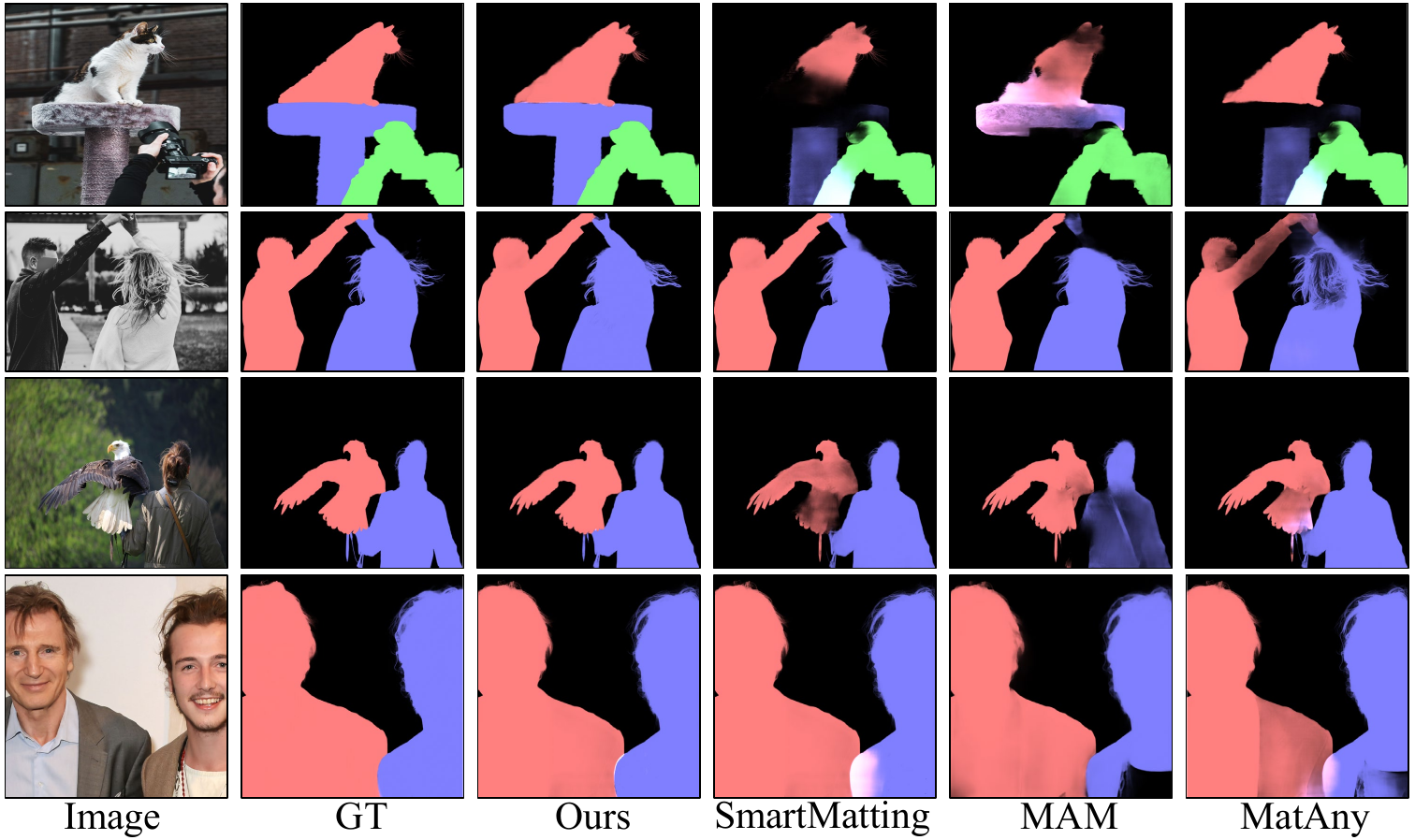}}
		\caption{Visual comparison with on HIM2K and RefMatte real-world multi-objects datasets. Note that ours is guided by text, while others are used by visual prompt (box).}
		\label{fig_matte}
	\end{figure}	
    \begin{figure*}[t]
		\centerline{\includegraphics[width=1.94\columnwidth]{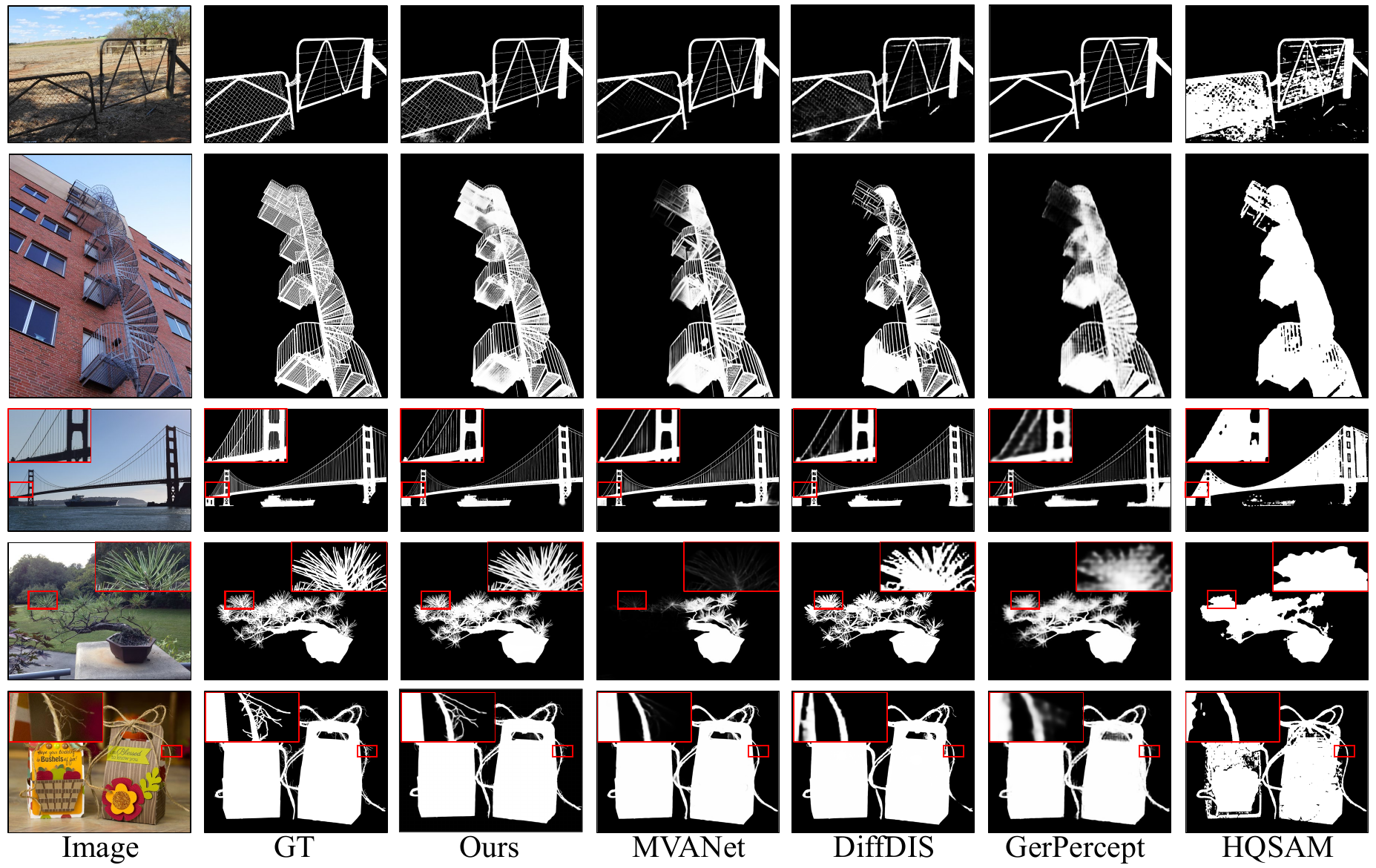}}
		\caption{Visual comparisons with different methods on DIS-5K dataset. Zoom in for the best view.}
		\label{fig:DIS}
	\end{figure*}
	\subsection{Optimization Objective}
		Here, we define the overall objective loss used to optimize our network. During knowledge distillation, we employ Kullback-Leibler (KL) divergence to measure the feature distance. In the domain alignment process, we incorporate Eq.~\ref{loss:adv} and Eq.~\ref{loss:ot} to constrain foreground learning. Considering the differences between tasks, we apply distinct prediction losses $\mathcal{L}_{\text{head}}$ for matting and segmentation. For the matting task, we follow ViTMatte~\cite{yao2024vitmatte} and adopt the $l_1$ loss together with the Laplacian loss. In contrast, for dichotomous segmentation, we employ the commonly used combination of BCE loss and IoU loss~\cite{yu2024multi}.
		The total term of our optimization objective is:
	\begin{equation}
		\mathcal{L} = \mathcal{L}_{\texttt{kd}} + \mathcal{L}_{\texttt{adv}} +\mathcal{L}_{OT}+\mathcal{L}_{\texttt{head}}.
	\end{equation}
	\section{Experiments}

	\subsection{Setups}
	\label{sec:exp}
	\subsubsection{Datasets and Metrics.}
	We evaluate our method on \textit{\textbf{real-world}} datasets. Only HIM2K~\cite{sun2022human} and RefMatte~\cite{li2023referring} support \textit{\textbf{multiple object instances}} matting configuration, and we use them to assess the referring matting capability of our approach. For other datasets, we choose DIS-5K~\cite{qin2022highly} as a representative for comparison here because it has a sufficient number of samples and its scenes are more complex and challenging. Results on other datasets can be found in our supplementary material due to space constraints. Since our pipeline requires paired images (sharing the same foreground) as inputs, we use the images from BG20K~\cite{li2022bridging} to replace the background yielding synthesized training images. Notably, we use the text prompt template \texttt{"a photo of \{CLS\}."} for the referring setting, except for RefMatte, which has provided text annotations.

    %We also examine the overall effectiveness of our design on DIS-5K. As HIM2 and DIS-5K do not provide text descriptions, we manually annotate them. Notably, we apply the prompt template \texttt{"a photo of \{CLS\}."} for DIS-5K, since it contains only a single object per mask. For other 
	
	Following the evaluation protocol established in HIM2K, we adopt the Instance Matting Quality (IMQ) metric to comprehensively assess matting performance. For RefMatte, we report results using a set of widely metrics, including Sum of Absolute Differences (SAD), Mean Squared Error (MSE), Gradient Distortion (Grad), and Connectivity Loss (Conn), which collectively provide insights into both global and local quality of the predicted alpha mattes. In the case of dichotomous segmentation, we follow the evaluation protocol proposed in~\cite{qin2022highly}, and employ several commonly used criteria such as the maximum F-measure ($maxF_\beta$), weighted F-measure ($F_\beta^w$), mean absolute error ($M$), and structural similarity index ($S_\alpha$).
	%%
	%We present more results on other datasets as well, which can be found in our \textit{Supplementary Material}.

	%We first extend the DIS5K dataset with text description. Following 

	\subsubsection{Implement details.} All the experiments are conducted via 4$\times$Nvidia 4090 GPUs. We utilize the AdamW optimizer with a constant learning rate of 1e-5 to train the model. We deploy Depth-Anything V2~\cite{yang2024depthv2} Large as teacher model. For a fair comparison, DINOV2~\cite{oquab2023dinov2} Base and Small are as student models for dichotomous segmentation and image matting, respectively. CLIP-Base~\cite{radford2021learning} is deployed for embedding text prompts.
	The threshold $\delta$ in DAD is 0.25, and the meta-net $\tau$ is built with a simple MLP structure (Linear-ReLU-Linear) for efficiency.
    We take the average performance as our final result with three runs by different seeds. For more details, please refer to our supplementary material.
	
	 %Following previous protocol, we xxx.
	 %For Matting task, we use xxx loss. In terms of 
	
\begin{table*}[t]
    \centering
    %\vspace{-2 em}
    \renewcommand\arraystretch{1.1}
    \setlength{\tabcolsep}{0.65mm}
    \resizebox{\textwidth}{!}{
        \begin{tabular}{r|c|cccccc|cccccc}
            \toprule[1.1pt]
            \multicolumn{2}{c|}{\multirow{3}{*}{\textbf{Configuration}}} & \multicolumn{6}{c|}{\textbf{Dataset: DIS-VD}} & \multicolumn{6}{c}{\textbf{Dataset: Overall DIS-TE (1-4)}}\\
            \cline{3-14}
            \multicolumn{2}{c|}{} & \multicolumn{6}{c|}{\textbf{Metric}} & \multicolumn{6}{c}{\textbf{Metric}} \\
            \midrule[1.1pt]
            \textbf{Methods}& \textbf{Venues}& $maxF_\beta\uparrow$ & $F^w_\beta\uparrow$ & $~M\downarrow$ & $S_\alpha\uparrow$ & $E_\phi^m\uparrow$ & $\HCE_\gamma\downarrow$ & $maxF_\beta\uparrow$ & $F^w_\beta\uparrow$ & $~M\downarrow$ & $S_\alpha\uparrow$ & $E_\phi^m\uparrow$ & $\HCE_\gamma\downarrow$ \\
            \midrule[1.1pt]
            IS-Net~\cite{qin2022highly} &ECCV$_{22}$& 0.791 & 0.717 & 0.074 & 0.813 & 0.856 & 1116 & 0.799 & 0.726 & 0.070 & 0.819 & 0.858 & 1016 \\
            PGNet~\cite{xie2022pyramid} &CVPR$_{22}$&0.798&0.733&0.067&0.824&0.879&/&0.809&0.746&0.063&0.830&0.885&/\\
            FP-DIS~\cite{zhou2023dichotomous} &IJCAI$_{23}$&0.823& 0.763& 0.062& 0.843& 0.891&1309&0.831& 0.770 &0.057 &0.847 &0.895&1164\\
            HitNet~\cite{hu2023high}&AAAI$_{23}$&0.805&0.757&0.061&0.828&0.890&1550&0.815&0.767&0.057&0.836&0.894&1001\\
            UDUN~\cite{pei2023unite}&MM$_{23}$&0.823&0.763&0.059&0.838&0.892&1097&0.831&0.772&0.057&0.844&0.892&977\\
            HQ-SAM~\cite{ke2023segment}&NIPS$_{23}$&0.739&0.706&0.120&0.740&0.818&1553&0.830&0.804&0.618&0.835&0.902&1299\\
            Pi-SAM~\cite{liu2024segment}&MM$_{24}$&0.883&0.866&0.035&0.889&\underline{0.945} &1322&0.890&0.873&0.033&0.893&\textbf{0.948} &1191\\
            BiRefNet~\cite{BiRefNet}&AIR$_{24}$&0.891&0.854&0.038&0.898&0.931&989&0.896&0.858&0.035&0.901&0.934&916\\
            FSANet~\cite{jiang2024high}&TNNLS$_{24}$&0.825&0.776&0.057&0.845&0.894&1094&0.841&0.794&0.052&0.857&0.903&973\\
            %Grounded-SAM~\cite{ren2024grounded}&Arxiv$_{24}$&\texttt{Text}&&&&&&&&&&&&\\
            MVANet~\cite{yu2024multi} &CVPR$_{24}$&0.904& 0.861 & 0.035 & \textbf{0.909} & 0.937 & \underline{878} & 0.916 & 0.855 & 0.035 & 0.905 & 0.938 &\underline{790} \\
            %LiSA~\cite{lai2024lisa}&CVPR$_{24}$&&&&&&&&&&&&\\
            %MAM~\cite{li2024matting}&CVPRW$_{24}$&&&&&&&&&&&&\\
            %SmartMatting~\cite{ye2024unifying}&CVPR$_{24}$&0.843&0.829&0.057&0.866&0.900&1086&0.876&0.810&0.065&0.874&0.907&1224\\
            MaskFactory~\cite{qian2024maskfactory}&NIPS$_{24}$&0.835&0.759&0.072&0.866&0.923&$/$&0.839&0.773&0.069&0.859& 0.907&$/$\\
            DiffDIS~\cite{yu2024high}&NIPS$_{24}$&\underline{0.918}&\underline{0.888}&\underline{0.029}&\underline{0.904} &\textbf{0.948}&$/$&\underline{0.921}&\underline{0.892}&\underline{0.028}&\underline{0.905}&\underline{0.947}&$/$\\
            Gerpercept~\cite{xu2024matters}&ICLR$_{25}$&0.877 &0.859& 0.035 &0.887& 0.941 &1262& 0.875 &0.856& 0.036 &0.885 &0.939& 1176\\
            %Text4Seg~\cite{lan2024text4seg}&ICLR$_{25}$&\texttt{Text}&&&&&&&&&&&&\\
            %Ours&$/$&& & & & & & &&&&&\\
           % \hdashline
           \midrule
            Ours&$/$&\textbf{0.924} &\textbf{0.900} &\textbf{0.025} &\textbf{0.909} &\underline{0.945} & \textbf{796} & \textbf{0.922} & \textbf{0.895} & \textbf{0.026} & \textbf{0.920} & \underline{0.947} & \textbf{580}\\
            \bottomrule[1.1pt]
        \end{tabular}
    }
        \caption{Quantitative comparisons on DIS5K~\cite{qin2022highly} validation and testing sets. The best performing results are highlighted with \textbf{bold} and runner-up results are \underline{underlined}.}
    \label{tab:dis}
\end{table*}

\subsection{Main Results}

	\subsubsection{Image Matting.} As shown in Tab.~\ref{tab:matte}, our method achieves the best results across almost all metrics on the image matting datasets. On \textsc{HIM2K-Natural}, we outperform the runner-up (SmartMatting~\cite{ye2024unifying}) by a notable margin in all four IMQ-based metrics, setting new records with 83.48 (IMQ-MSE), 73.45 (IMQ-MAD), 75.69 (IMQ-Grad), and 74.83 (IMQ-Conn). Similarly, on \textsc{RefMatte-RW}, our method obtains the lowest error in MSE (0.010), SAD (21.31), and Gradient (19.38), while achieving competitive performance in Connectivity, ranking second. 

	We present the visual comparisons in Fig.~\ref{fig_matte}. Our method produces more accurate and detailed results compared to existing approaches. These improvements are particularly evident in challenging cases with complex object boundaries and ambiguous foreground-background transitions.

	\subsubsection{Accurate Dichotomous Segmentation.} Our framework is also applicable for DIS5K. As shown in Tab.~\ref{tab:dis}, on the DIS-VD split, our method not only surpasses the previous \sota by a clear margin in key metrics such as $maxF_\beta$ and $M$, but also places among the top two overall, often trailing only slightly behind the second-best performer in those where it does not lead. A similar trend is observed on the more comprehensive DIS-TE, where we also obtain the top performance in $maxF_\beta$ (0.922), $F^w_\beta$ (0.895), $M$ (0.026), and $S_\alpha$ (0.920), with competitive results in $E_\phi^m$ (0.947) and the lowest $\HCE_\gamma$ (580).
	
	The qualitative comparisons can be found in Fig.~\ref{fig:DIS}. Our method achieves superior performance in handling complex real-world images, particularly in preserving fine details and maintaining accurate boundaries. For example, the ropes on the viaduct (3rd row) and cilia on the rope (last row).

	\subsection{Ablation Studies}
	Unless specification, we conduct the ablation studies on RefMatte~\cite{li2023referring} dataset. 
	
		\begin{table}[t]
		\centering
		\resizebox{0.47\textwidth}{!}{
			\setlength{\tabcolsep}{0.4mm}
			\begin{tabular}{c|cccc}
				\Xhline{1.02pt}
				\multicolumn{1}{c|} {Methods} &\multicolumn{1}{c}{MSE$\downarrow$} &\multicolumn{1}{c}{SAD$\downarrow$}&\multicolumn{1}{c}{Grad$\downarrow$}&\multicolumn{1}{c}{Conn$\downarrow$}\\
				\hline
				w/o KD &0.018&26.34&23.48&4.38\\
				Vanilla KD&0.014&23.48&22.84&3.99\\
					$\tau_{+}^{A}$ &0.016&23.40&23.02&3.90\\ 
					$\tau_{+}^{B}$ &0.014&22.84&21.92&3.88\\ 
				$\tau_{+}^{A}+\tau_{+}^{B}$ &0.013&22.18&20.29&3.60\\ 
$\tau_{+}^{A}+\tau_{+}^{B}+\tau_{-}^{A}+\tau_{-}^{B}$ (Ours) &\textbf{0.010}&\textbf{21.31}&\textbf{19.38}&\textbf{3.33}\\
				\Xhline{1.02pt}
			\end{tabular}
			}
		\caption{Ablation study on the distillation.}
        \label{tab:exp1}
	\end{table}
	\subsubsection{Effectiveness of Distillation.} We conduct an ablation study on our knowledge distillation strategy, as shown in Tab.~\ref{tab:exp1}. As shown in the table, omitting knowledge distillation (w/o KD) leads to the poorest performance. Incorporating standard knowledge distillation without depth-aware guidance (Vanilla KD) only yields a marginal improvement. By introducing the meta-net, distilling knowledge from the foreground features of either image A or B individually improves model results, demonstrating the effectiveness
	 of feature-specific distillation. Combining both foreground distillations yields further gains. Our final model extends this strategy by also distilling background, leading to the optimal configuration.
\begin{table}[t]
 \resizebox{\linewidth}{!}{
			\setlength{\tabcolsep}{2mm}
			\begin{tabular}{c|cccc}
				\Xhline{1.02pt}
				\multicolumn{1}{c|}{Methods} & \multicolumn{1}{c}{MSE$\downarrow$} & \multicolumn{1}{c}{SAD$\downarrow$} & \multicolumn{1}{c}{Grad$\downarrow$} & \multicolumn{1}{c}{Conn$\downarrow$} \\
				\hline
				w/o alignment &0.157&24.33&26.87&6.07\\
				$\mathcal{L}_{adv}$ &0.128&22.79&23.01&4.83 \\
				$\mathcal{L}_{OT}$&0.134&22.47&20.72&4.34\\
				%\rowcolor{gray!10}
				$\mathcal{L}_{adv}$+$\mathcal{L}_{OT}$ (Ours) &\textbf{0.010}&\textbf{21.31}&\textbf{19.38}&\textbf{3.33}\\
				\Xhline{1.02pt}
			\end{tabular}
		}
		\caption{Ablation study on the loss of domain alignment.}
		\label{tab:exp2_left}
\end{table}

\begin{table}[t]
\resizebox{\linewidth}{!}{
			\setlength{\tabcolsep}{2.5mm}
			\begin{tabular}{c|cccc}
				\Xhline{1.02pt}
				\multicolumn{1}{c|}{Methods} & \multicolumn{1}{c}{MSE$\downarrow$} & \multicolumn{1}{c}{SAD$\downarrow$} & \multicolumn{1}{c}{Grad$\downarrow$} & \multicolumn{1}{c}{Conn$\downarrow$} \\
				\hline
				point &0.009&18.46&15.44&3.01\\
				box & \textbf{0.006} & \textbf{16.37} & \textbf{13.89} & \textbf{2.48}\\
				%\rowcolor{gray!10}
				text (Ours) &0.010&21.31&19.38&3.33\\
				\Xhline{1.02pt}
			\end{tabular}
		}
		\caption{Ablation study on the prompt types.}
		\label{tab:exp2_right}
\end{table}

	\begin{figure}
		\centering
\includegraphics[width=0.85\linewidth]{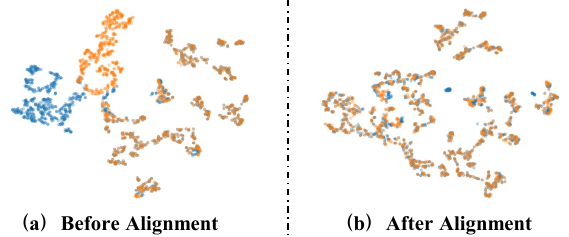}
		\caption{t-SNE visualization from the foreground tokens.}
	\label{fig:tsne}
	\end{figure} 
	\subsubsection{Effectiveness of Domain Adaptation.}
	We evaluate the impact of different alignment losses, as shown in Tab.~\ref{tab:exp2_left}. Removing alignment entirely (w/o alignment) leads to the worst performance, highlighting its importance. While $\mathcal{L}_{adv}$ improves global structure (lower MSE/SAD), $\mathcal{L}_{OT}$ better preserves edge details and foreground connectivity (lower Grad/Conn). Combining both achieves the best results, leveraging their complementary strengths.
	Moreover, Fig.\ref{fig:tsne} illustrates the effect of our alignment strategy using t-SNE visualizations. In (a) "Before Alignment", the feature clusters are distinctly separated and highly scattered, indicating poor consistency between different semantic regions. After applying our alignment loss, the feature clusters become more compact and exhibit significant overlap, demonstrating improved coherence and discriminability. This qualitative improvement aligns with the quantitative results in Tab.~\ref{tab:exp2_left}, where removing alignment leads to a sharp performance drop across all metrics.

	\subsubsection{Prompt Probing.} We also evaluate the effectiveness of different prompts. As shown in Tab.~\ref{tab:exp2_right}, visual prompts yield better results compared to text-based guidance, likely due to their stronger spatial constraints, which help more accurately localize the foreground object. However, these visual prompts are class-agnostic, limiting their ability to distinguish between semantically meaningful regions. Therefore, we report the results based on text prompts.
	
	%In contrast, text-based prompts, although more flexible and user-friendly, result in slightly lower accuracy, suggesting room for improvement in semantic-to-matte alignment. These results indicate that while precise spatial cues (e.g., box) lead to better quantitative performance, our framework also supports diverse interaction modes with acceptable trade-offs.

	%\section{Limitation and Future Work}
	%\label{sec:limitation}
	%Despite the effectiveness of FCLM, our method still has several limitations. First, while supporting both visual and text prompts, the performance of text-driven prediction lags behind that of visual inputs, indicating a need for better vision-language alignment. Second, our domain-invariant learning reduces reliance on synthetic data but does not fully eliminate it. In the future, we aim to improve cross-modal fusion and explore strategies based on the prevailing MLLMs. Collecting high-quality real-world datasets also remains an important direction for further improvement.
	
	\section{Conclusions}
	In this work, we introduce a Foreground Consistent Learning Model (FCLM) for high-accuracy image matting and segmentation toward real-world. Motivated by the limitations of current datasets, we formulate the use of synthetic training data as a domain adaptation problem and introduce distillation and alignment strategies to more effectively learn discriminative foreground features. To accommodate a variety of interaction modes, our network supports both visual and textual prompts, facilitating semantic-aware and target-specific predictions. Our method achieves \sota results on publicly accessible real-world datasets, validating its effectiveness in practical scenarios.
    
%\section{Acknowledgments}
%This work is supported by OPPO Research Fund, the Guangdong Science and Technology Department (No. 2024ZDZX2004), the Guangzhou Industrial Information and Intelligent Key Laboratory Project (No. 2024A03J0628), the China Southern Power Grid Science and Technology Project (Project No.: 030117KC23120005), and the Nansha Key Area Science and Technology Project (No. 2024ZD006).
\clearpage
\bibliography{aaai2026}

\end{document}